\documentclass[11pt]{article}
\usepackage{booktabs}
\usepackage{multirow}
\usepackage{xcolor}
\usepackage{graphicx}
\usepackage{subcaption}
\usepackage{capt-of}
\usepackage[normalem]{ulem}
\usepackage{todonotes}
\usepackage{hyperref}
\usepackage{float}
\usepackage{placeins}
\usepackage{cuted}

\usepackage{acl}

\usepackage{times}
\usepackage{latexsym}
\usepackage{amsmath}
\usepackage[T1]{fontenc}

\usepackage[utf8]{inputenc}
\usepackage{booktabs}

\usepackage{microtype}

\usepackage{inconsolata}

\title{Loss-Based Active Learning for Neural Abstractive Summarization}

\author{
    Michail Ioannou,
    Tatiana Passali,
    George Michalopoulos,
    Grigorios Tsoumakas \\
    School of Informatics \\
    Aristotle University of Thessaloniki \\
    Greece \\
    \texttt{mioannoug@csd.auth.gr, scpassali@csd.auth.gr,}\\
    \texttt{georgemi@microsoft.com, greg@csd.auth.gr}
}

\begin{document}

\maketitle
\begin{abstract}
Fine-tuning abstractive summarization models requires high-quality annotated data. However, obtaining such corpora is expensive and time-consuming, as it requires human annotators to read and comprehend long documents to create accurate summaries. Active learning mitigates this issue by selecting only the most informative instances for annotation, allowing models to achieve competitive results with significantly fewer labels. However, the application of active learning to summarization remains under-explored, and existing studies often suffer from instability and significant computational bottlenecks. To overcome these challenges, we propose LOBSTER (LOss-BaSed acTivE leaRning), a novel active learning framework designed specifically for abstractive summarization. LOBSTER improves performance by prioritizing unlabeled instances semantically similar to the model's current high-loss training examples, enabling the model to explicitly correct its specific weaknesses. Our empirical evaluation across three benchmark datasets and two summarization backbone models demonstrates that LOBSTER consistently matches or outperforms current state-of-the-art approaches while achieving a query selection speedup of up to $665\times$.
\end{abstract}

\section{Introduction}
Abstractive summarization generates concise summaries by producing novel words and phrases rather than directly copying from the source \citep{see-etal-2017-get, pmlr-v119-zhang20ae}. State-of-the-art performance has been achieved by pre-trained large language models (LLMs) \citep{lewis-etal-2020-bart, pmlr-v119-zhang20ae, zaheer2020bigbird}, based on the Transformers architecture \citep{vaswani2017attention}. Nevertheless, fine-tuning LLMs for abstractive summarization is costly, since producing high-quality summaries requires human annotators to carefully read the full document and identify important information \citep{tsvigun-etal-2022-active, perlitz-etal-2023-active}.

Active learning addresses this challenge by iteratively selecting the most informative examples for annotation. This enables models to achieve strong performance with fewer training instances, thereby reducing the overall cost of obtaining labeled data \citep{settles2012active}. While recent efforts to apply active learning to abstractive summarization have shown promise, existing methods suffer from critical limitations. Uncertainty-based strategies are computationally demanding and 
tend to select outliers, as these instances often exhibit high uncertainty \citep{GIDIOTIS2024101553}. Conversely, diversity-based approaches can be unstable since they may overemphasize dense regions of the data space \citep{tsvigun-etal-2022-active}, and more recent methods rely on external LLM systems, which can introduce additional cost and dependency \citep{li-etal-2024-active}.

This work introduces a novel technique that strikes a balance between summarization performance and computational efficiency. We demonstrate that the cross-entropy loss of current labeled examples provides a useful signal for guiding the model toward the specific data it requires to improve. Building on this insight, we present LOBSTER (LOss-BaSed acTivE leaRning), a method that integrates density-based diversity with a mechanism that projects the hardness of labeled examples to unlabeled ones based on their semantic similarity. Our experiments demonstrate that LOBSTER matches or surpasses state-of-the-art performance while being significantly faster than uncertainty-based strategies and more stable than methods relying solely on diversity. To the best of our knowledge, we are the first to use sequence-to-sequence cross-entropy loss of labeled data to guide the  selection of unlabeled data in abstractive summarization. 

The main contributions of this work are as follows: 
\begin{itemize}
    \item We propose a hybrid selection strategy that explicitly targets the weaknesses of the current model while maintaining data diversity.
    \item We conduct extensive experiments across three benchmark datasets and two model backbones, demonstrating the effectiveness of LOBSTER while highlighting the inherent instability of diversity-based methods and the high latency of uncertainty-based strategies.
    \item We present a detailed data efficiency analysis revealing the surprising competitiveness of random sampling, under large annotation budgets.
\end{itemize}

\section{Related Work}
\label{sec:related_work}

\subsection{Abstractive Summarization} 
The introduction of neural sequence-to-sequence models \citep{sutskever2014sequence} and the attention mechanism \citep{bahdanau2015neural} established the foundations for modern abstractive summarization. Building on these advances, the Transformer architecture \citep{vaswani2017attention} enabled large-scale pre-trained language models such as BART~\citep{lewis-etal-2020-bart}, T5~\citep{JMLR:v21:20-074} and PEGASUS~\citep{pmlr-v119-zhang20ae}, which set the state of the art. Recent work introduced model adaptations for long or multi-document inputs \citep{khanna-etal-2022-transformer,xiao-etal-2022-primera}, while instruction-tuned LLMs demonstrated strong few-shot and zero-shot capabilities across various tasks, including abstractive summarization, without task-specific fine-tuning \citep{achiam2023gpt,JMLR:v25:23-0870}.

\subsection{Active Natural Language Generation}
Within the field of NLP, active learning has been extensively studied and applied to discriminative tasks such as text classification \citep{ein-dor-etal-2020-active,margatina-etal-2021-active,yu-etal-2022-actune} and sequence tagging \citep{shen-etal-2017-deep,radmard-etal-2021-subsequence}. However, its application to generative tasks has received comparatively little attention \citep{perlitz-etal-2023-active}. Existing active learning approaches for natural language generation are mainly categorized into uncertainty-based, diversity-based, and hybrid methods \citep{zhang-etal-2022-survey}.

\paragraph{Uncertainty-based methods.} These methods select examples for annotation where the model is least confident in its predictions \citep{settles2012active}. \citet{gidiotis-tsoumakas-2022-trust} leverage Bayesian approximation to model uncertainty, by adapting the Monte Carlo BLEU variance metric \citep{xiao2020watzeijedetecting} to summarization. They subsequently apply this uncertainty estimation framework to active learning via their Bayesian Active Summarization (BAS) approach \citep{GIDIOTIS2024101553}. Documents with the highest variance are selected for annotation, and the model is iteratively retrained on the labeled set. Empirical results indicate that this approach achieves performance comparable to full-data training with only a few hundred examples under specific experimental settings. Despite these gains, however, purely uncertainty-based strategies in abstractive summarization can be sensitive to data quality, often remaining susceptible to selecting noisy or uninformative examples \citep{tsvigun-etal-2022-active, li-etal-2024-active}.

\paragraph{Diversity-based methods.} The goal of these methods is to select examples that are both diverse with respect to the labeled set while being representative of the data distribution \citep{kim-etal-2006-mmr}. \citet{tsvigun-etal-2022-active} propose In-Domain Diversity Sampling (IDDS), which selects instances that differ from the labeled set while still remaining close to the unlabeled pool, balancing diversity with domain relevance. Building on this idea, \citet{li-etal-2024-active} introduce LLM-Determined Curriculum Active Learning (LDCAL), which leverages LLMs to rate instance difficulty and guide training. They also propose {\em certainty gain maximization}, a complementary strategy that selects instances whose distribution aligns closely with the overall data.

\paragraph{Hybrid methods.} Recent work has shown that combining uncertainty and diversity methods, can outperform uncertainty-only and diversity-only methods in multiple settings \citep{azeemi-etal-2025-label,giouroukis2025dualdiversityuncertaintyactive}. \citet{azeemi-etal-2025-label} introduce Hybrid Uncertainty and Diversity Sampling (HUDS), a hybrid strategy for neural machine translation that combines normalized negative log-likelihood for uncertainty with cosine distance from cluster centroids for diversity. For abstractive summarization, Diversity and Uncertainty Active
Learning (DUAL) \citep{giouroukis2025dualdiversityuncertaintyactive} first selects a compact, diverse yet representative subset via IDDS, and then measures uncertainty within that subset using BAS.

\section{Methodology}
Under cross-entropy training, high-loss examples tend to generate larger gradients and thus dominate parameter updates, a property reflected in hard example mining~\cite{shrivastava2016training} and boosting methods~\cite{freund1997decision}. We exploit this property in the active learning setting by directing supervision toward unlabeled instances that are semantically aligned with high-loss training samples, concentrating annotation budget in regions where parameter updates have greater impact. Our core hypothesis is that an active learning strategy can improve performance by prioritizing unlabeled instances that are semantically similar to the model's current high-loss training samples. By explicitly targeting the unlabeled \textit{semantic counterparts} of these hard examples, the model is driven to correct its specific weaknesses.

\begin{figure*}[t]
    \centering
    \includegraphics[width=0.98\textwidth]{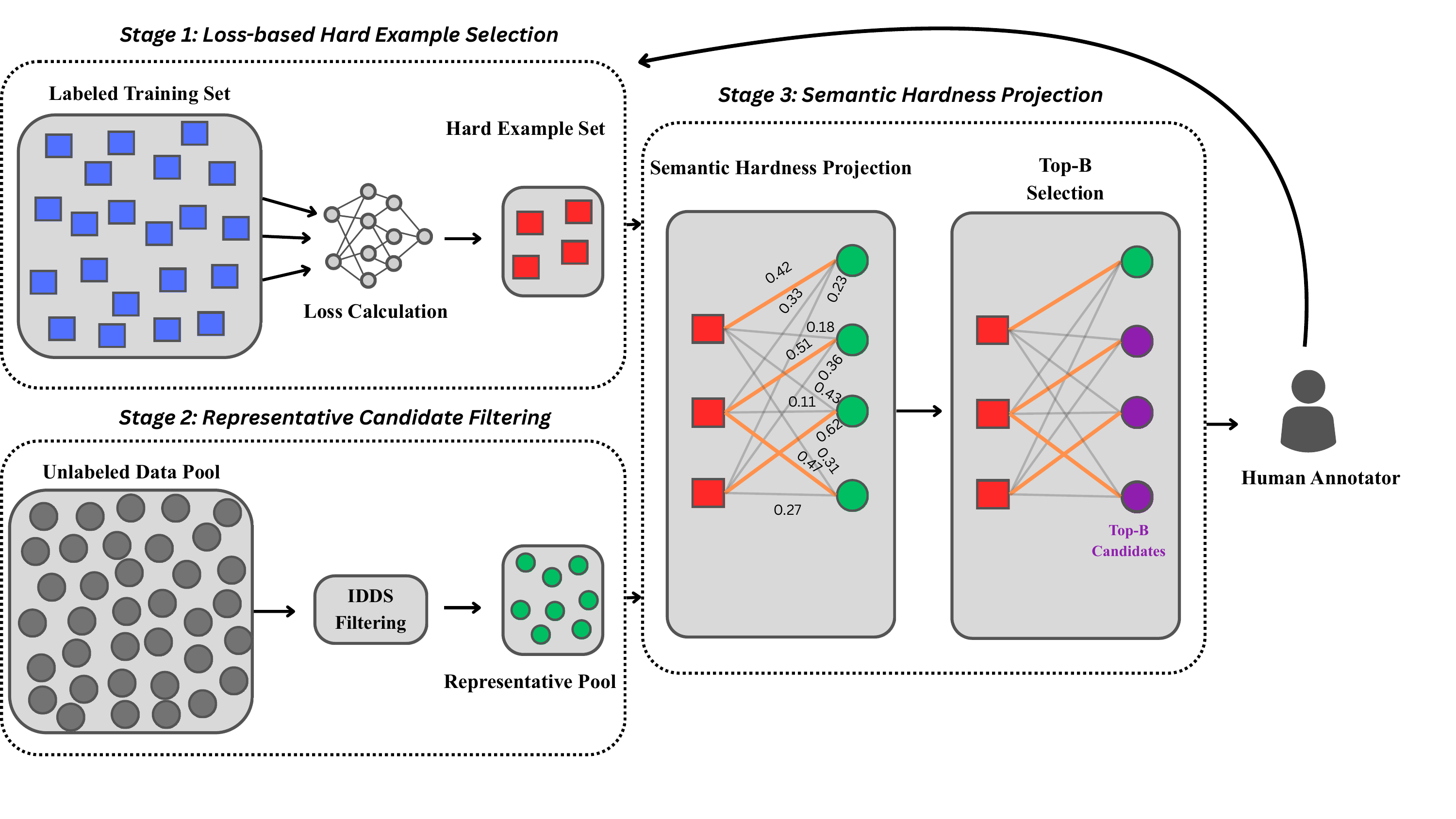}
    \caption{The three-stage pipeline of LOBSTER.
    (1) \textbf{Loss-based Hard Example Selection} identifies ``Hard Example'' set $\mathcal{H}$ (red squares) via high cross-entropy loss to serve as semantic anchors. 
    (2) \textbf{Representative Candidate Filtering} uses IDDS to distill a pool of representative candidates (green circles) from the unlabeled set $\mathcal{U}$. 
    (3) \textbf{Semantic Hardness Projection} computes the Max-Similarity Score $S(x_u)$ (Eq. 3) using Sentence-BERT embeddings. This helps to identify a ``semantic twin'' for each hard example, forming the final Top-$B$ batch for annotation.}
    \label{fig:methodology}
\end{figure*}

To achieve this, we introduce LOBSTER (LOss-BaSed acTivE leaRning), a 3-stage sample acquisition strategy that first identifies high-loss examples in the labeled set, then applies density-based filtering to construct a representative candidate pool, and finally selects samples that are semantically similar to the difficult instances without the need for the expensive autoregressive decoding used by uncertainty methods. The complete workflow of the LOBSTER strategy is illustrated in Figure \ref{fig:methodology}.

\subsection{Problem Formulation}
We define the active learning task for abstractive text summarization as follows. Let $\mathcal{U}$ represent the large unlabeled pool of documents, and $\mathcal{L}$ represent the initial small labeled set. At each active learning cycle $t$, an acquisition function utilizes the current model $\theta_t$ to select a batch of $B$ instances from $\mathcal{U}$ to be annotated. The goal is to maximize summarization performance on a held-out test set while minimizing the total number of annotated samples.

\subsection{Loss-Based Active Learning}
LOBSTER is a hybrid active learning method that combines loss-guided sampling with IDDS in order to maximize performance. It consists of three stages which we introduce below.

\paragraph{Stage 1: Loss-Based Hard Example Selection.}
Unlike classification tasks, where errors correspond to strict binary failures, difficulty in abstractive summarization is better reflected by generation loss. High-loss examples are therefore more likely to correspond to error-prone regions of the data distribution. The first stage of LOBSTER aims to diagnose these regions. Unlike uncertainty sampling, which estimates uncertainty on unlabeled data, we explicitly measure generation difficulty on the \textit{labeled} data. We first compute the token-level cross-entropy loss for all samples in the current labeled set $\mathcal{L}$. For a given source document $x$ and target summary $y$ of length $T$, the loss is defined as the average negative log-likelihood over the sequence:
\begin{equation}
    L(x, y) = - \frac{1}{T} \sum_{t=1}^{T} \log P(y_t \mid y_{<t}, x; \theta)
\end{equation}
We then identify the set of {\em hard examples}, $\mathcal{H}$, defined as the top-$k$ examples with the highest loss:
\begin{equation}
    \mathcal{H} = \{x_i \in \mathcal{L} \mid L(x_i, y_i) \geq \tau_k \}
\end{equation}
where $\tau_k$ corresponds to the loss of the $k$-th example in $\mathcal{L}$ when sorted by loss in descending order. This acts as a dynamic threshold that adapts to the model's current performance, rather than a fixed predefined threshold.
These high-loss instances serve as \textit{semantic anchors}, representing the specific linguistic patterns or domains where the model is currently underperforming.

\paragraph{Stage 2: Representative Candidate Filtering.}
\label{sec:stage2}
We observed that retrieving candidates purely via embedding similarity to the hard example set frequently produced clusters of highly correlated, near-duplicate examples. Therefore, we employ IDDS \citep{tsvigun-etal-2022-active} as a diversity-aware pre-filter to ensure the candidate pool covers broad, representative regions of the data distribution. Crucially, this mitigates the risk of selecting noisy outliers (a common failure mode in uncertainty-based active learning \citep{tsvigun-etal-2022-active}). Furthermore, while density-based methods are known to oversample dense regions when used as a final selector \citep{li-etal-2024-active}, we avoid this issue by using IDDS strictly to construct a large candidate pool ($U_{rep} \gg B$) rather than for final budget allocation. Filtering is separated from the final selection in Stage 3, ensuring the candidate pool is representative and distinct from already labeled data, effectively mitigating the risks of both outlier selection and dense-region oversampling.

\paragraph{Stage 3: Semantic Hardness Projection.}
\label{sec:stage3}
The core contribution of LOBSTER is the explicit targeting of high-loss regions, i.e., difficult regions of the data. We assume that the optimal sample to annotate is the {\em semantic twin} of a known hard example. Under this assumption, documents that are semantically close to high-loss labeled examples are likely to provide informative supervision for improving the model.

For every document $x_u$ in the representative candidate pool $\mathcal{U}_{rep}$, we compute a maximum similarity score $S(x_u)$ against the set of hard examples $\mathcal{H}$ using Sentence-BERT \citep{reimers-gurevych-2019-sentence} embeddings $\phi(\cdot)$:
\begin{equation}
    S(x_u) = \max_{x_h \in \mathcal{H}} \left( \frac{\phi(x_u) \cdot \phi(x_h)}{\| \phi(x_u) \| \| \phi(x_h) \|} \right)
\end{equation}
Crucially, we use the $\max$ operator rather than the average. We do not want documents that are broadly related to the hard example set, rather we are searching for instances that are similar to specific hard examples. The final batch is constructed by selecting the top $B$ candidates with the highest $S(x_u)$.

\section{Experimental Setup}
\subsection{Datasets}
Following standard benchmarks in the field, we evaluated our active summarization models using three widely used datasets representing diverse summarization styles. AESLC contains short emails with their subject lines as summaries \citep{zhang-tetreault-2019-email}. XSum contains BBC news articles with their single-sentence summaries, requiring high-level abstraction \citep{narayan-etal-2018-dont}. Finally, the CNN/DailyMail dataset consists of news articles accompanied by multi-sentence, bullet-point summaries, providing a benchmark for longer-form abstractive summarization \citep{hermann2015teaching, nallapati-etal-2016-abstractive, see-etal-2017-get}. Table \ref{tab:dataset_stats} in Appendix~\ref{sec:appendix_dataset_statistics_and_hyperparameters} reports the size of the training and test sets, alongside the average token lengths for both source documents and reference summaries. All datasets consist entirely of English text. We cleaned all datasets by removing noisy instances, such as empty strings and duplicates. Additionally, we filtered the XSum and CNN/DailyMail datasets by excluding documents with fewer than 10 tokens and summaries with fewer than 3 tokens. Finally, due to the high computational cost of the uncertainty-based baselines across multiple active learning cycles, we evaluated our models on a random subset of 1,000 instances from the test sets.

\subsection{Active Learning Setup}
We follow the standard active learning framework used in prior studies \citep{siddhant-lipton-2018-deep,shelmanov-etal-2021-active}. We initialize each cycle with a randomly sampled seed set of 10 annotated instances, which enables model-dependent strategies to obtain initial acquisition scores. At each iteration, the top 10 examples from the unlabeled pool $\mathcal{U}$ are selected according to the acquisition strategy and added to the labeled pool $\mathcal{L}$ with their ground-truth summaries. We then fine-tune the summarization model from scratch on $\mathcal{L}$ and evaluate it on the held-out test set. The process is repeated for 15 iterations, resulting in a total annotation budget of 150 instances, and results are averaged over 5 independent runs.

\subsection{Baselines}
We compare our approach against the following baselines and state-of-the-art active summarization methods: random sampling, IDDS \citep{tsvigun-etal-2022-active}, BAS \citep{GIDIOTIS2024101553}, and DUAL \citep{giouroukis2025dualdiversityuncertaintyactive}. As discussed in Section \ref{sec:related_work}, these methods represent a diverse spectrum of strategies, including diversity-based, uncertainty-based, and hybrid approaches. To ensure a fair comparison with previous studies, we used the code provided by the respective authors.

\subsection{Implementation Details}
We use BART-base \citep{lewis-etal-2020-bart} and PEGASUS-large \citep{pmlr-v119-zhang20ae} as the summarization backbones. Hyperparameters were tuned on a small randomly sampled subset of the Gigaword dataset \citep{graff2003english}. All experiments were conducted using the HuggingFace Transformers library, with model-specific hyperparameters reported in Table~\ref{tab:hyperparameters} in Appendix~\ref{sec:appendix_dataset_statistics_and_hyperparameters}. All experiments were conducted on a Google Cloud Platform (GCP) virtual machine equipped with an NVIDIA L4 GPU. Our code is publicly available at \href{https://anonymous.4open.science/r/lobster-active-learning-397D}{GitHub}.

\subsection{Evaluation Metrics}
We evaluate summary quality using both lexical overlap and semantic similarity metrics. For lexical overlap, we compute ROUGE scores \citep{lin-2004-rouge}, including ROUGE-1, ROUGE-2, and ROUGE-L, which measure unigram overlap, bigram overlap, and longest common subsequence overlap, respectively. For semantic similarity, we additionally use BERTScore \citep{zhang2020bertscore}, which leverages pre-trained contextual embeddings to measure the alignment in meaning between generated and reference summaries. Although we report all ROUGE variants, our analysis primarily focuses on ROUGE-1 and BERTScore, as they provide complementary measures of lexical content preservation and semantic similarity. We also note that ROUGE-2 and ROUGE-L follow trends consistent with ROUGE-1 across all experiments.

\section{Results Analysis}
\subsection{Performance Comparison}
\label{sec:performance_comparison}
\begin{figure*}[t]
    \centering
    \renewcommand{\thesubfigure}{\Alph{subfigure}}
    
    \begin{subfigure}[t]{0.32\textwidth}
        \centering
        \includegraphics[width=\linewidth]{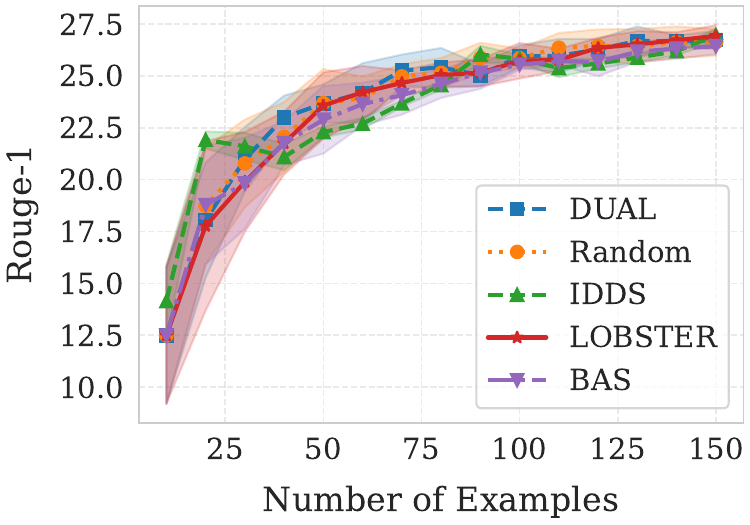}
        \caption{BART on AESLC}
        \label{fig:bart_aeslc}
    \end{subfigure}
    \hfill
    \begin{subfigure}[t]{0.32\textwidth}
        \centering
        \includegraphics[width=\linewidth]{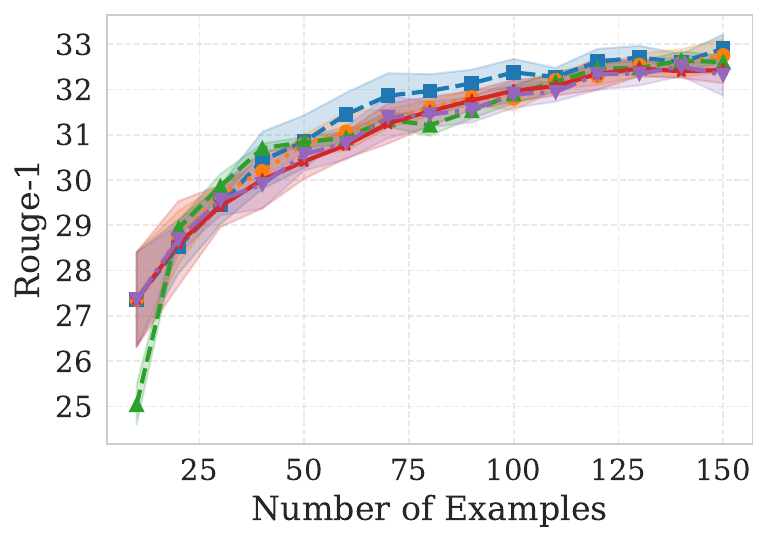}
        \caption{BART on XSum}
        \label{fig:bart_xsum}
    \end{subfigure}
    \hfill
    \begin{subfigure}[t]{0.32\textwidth}
        \centering
        \includegraphics[width=\linewidth]{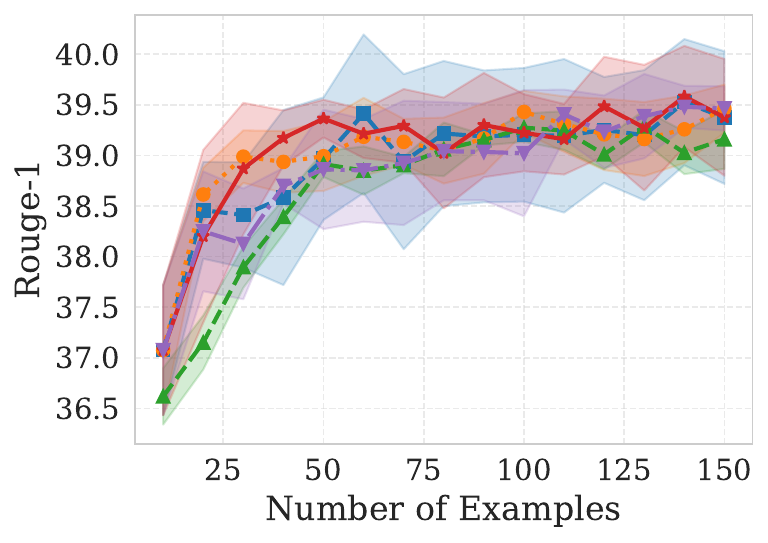}
        \caption{BART on CNN/DM}
        \label{fig:bart_cnn}
    \end{subfigure}

    \vspace{0.3cm} 

    \begin{subfigure}[t]{0.32\textwidth}
        \centering
        \includegraphics[width=\linewidth]{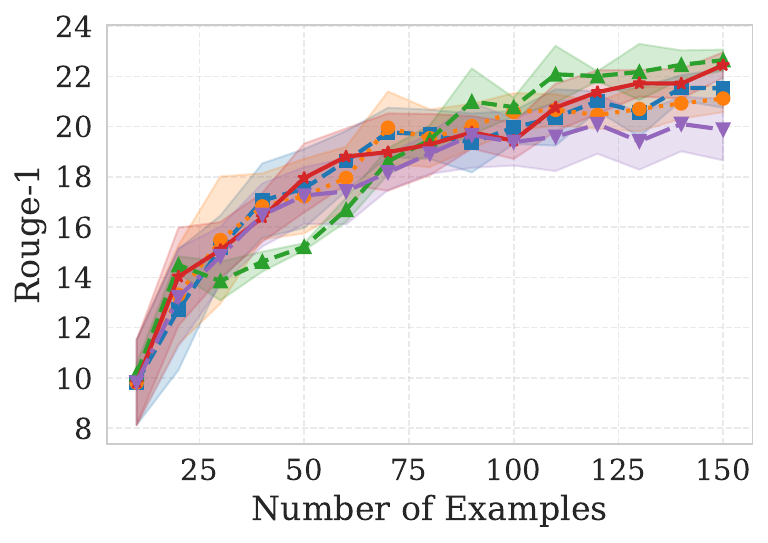}
        \caption{PEGASUS on AESLC}
        \label{fig:pegasus_aeslc}
    \end{subfigure}
    \hfill
    \begin{subfigure}[t]{0.32\textwidth}
        \centering
        \includegraphics[width=\linewidth]{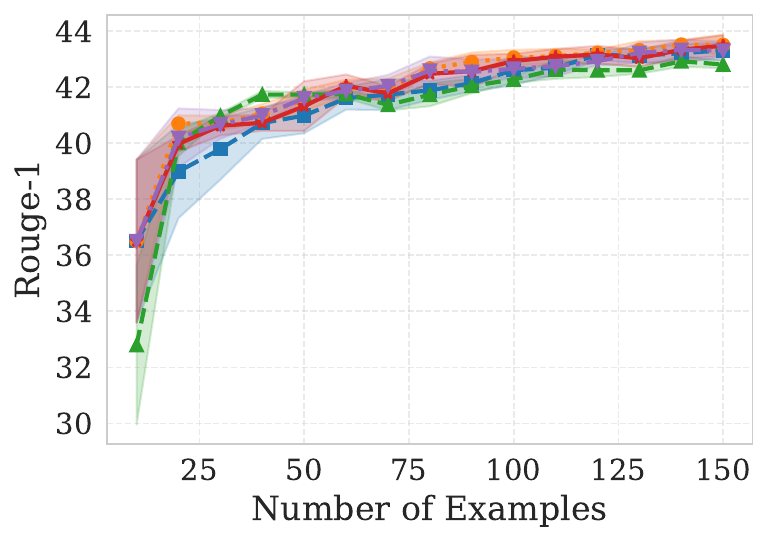}
        \caption{PEGASUS on XSum}
        \label{fig:pegasus_xsum}
    \end{subfigure}
    \hfill
    \begin{subfigure}[t]{0.32\textwidth}
        \centering
        \includegraphics[width=\linewidth]{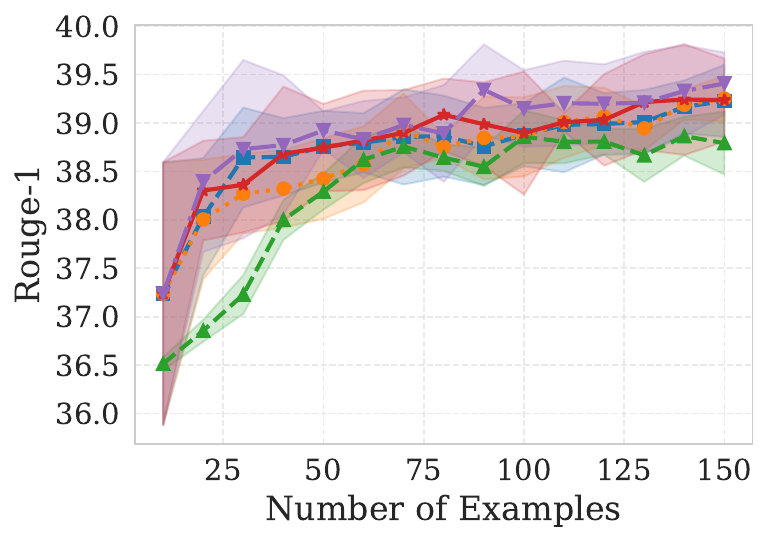}
        \caption{PEGASUS on CNN/DM}
        \label{fig:pegasus_cnn}
    \end{subfigure}

    \caption{Number of labeled instances vs. ROUGE-1. Comparison of active strategies using BART and PEGASUS backbones on AESLC, XSum, and CNN/DM datasets.}
    \label{fig:main_results_grid_rouge_1}
\end{figure*}

We evaluate the summarization performance of LOBSTER against four strategies across three benchmark datasets and two backbone architectures. Figures~\ref{fig:main_results_grid_rouge_1} and~\ref{fig:main_results_grid_bertscore} illustrate the ROUGE-1 and BERTScore learning curves, respectively, on the test set across active learning iterations, where solid lines represent the mean score across 5 random seeds and shaded regions indicate the standard deviation.
The full results at the final annotation budget are reported in Appendix~\ref{sec:appendix_summarization_performance}.

\paragraph{Competitive Effectiveness.}
Our analysis indicates that the efficacy of active learning strategies is often context-dependent, varying significantly across different datasets, metrics, and  architectures. For example, regarding ROUGE-1 scores on the CNN/DM dataset, BAS achieves the top performance with the PEGASUS model. Conversely, on the XSum dataset using the BART backbone, DUAL emerges as the strongest performer. These dataset- and model-specific variations are similarly observed when evaluating semantic quality with BERTScore. This variability suggests that standard active learning approaches are not universally optimal and are sensitive to the interaction between model architecture and summarization style. However, LOBSTER exhibits strong results, delivering performance that is consistently competitive with or superior to state-of-the-art baselines across these diverse experimental conditions. Specifically, on highly abstractive benchmarks such as XSum, LOBSTER achieves very strong scores with the PEGASUS backbone, outperforming or matching computationally expensive uncertainty-based approaches like BAS and DUAL. To further assess the reliability of these observations at the final annotation budget, we conducted paired permutation tests between LOBSTER and each baseline method across all datasets and backbone architectures. As reported in Table \ref{tab:statistical_tests}, several comparisons yield statistically significant differences ($p < 0.05$), indicating that the observed performance variations are statistically meaningful in specific experimental settings.

\paragraph{Addressing the Cold Start Problem.}
A key finding is LOBSTER's ability to correct the inherent weaknesses of the model. Although LOBSTER builds upon the IDDS framework, it demonstrates superior performance. While IDDS is generally competitive, it exhibits notable instability in our experiments, particularly on the CNN/DM dataset and on AESLC with PEGASUS. As seen in Figures ~\ref{fig:main_results_grid_rouge_1}C,~\ref{fig:main_results_grid_rouge_1}D, and~\ref{fig:main_results_grid_rouge_1}F and their corresponding BERTScore plots, IDDS suffers from a severe “cold start” problem, lagging significantly behind all other methods during the initial acquisition cycles. This initial instability is particularly critical because the core premise of active learning is to maximize performance under strict annotation budgets. Furthermore, the BERTScore evaluation confirms that this lag is not just a failure to predict exact n-grams, but a drop in actual semantic understanding. In contrast, LOBSTER demonstrates more stable performance throughout the active learning process, as it avoids the initial cold start problem that IDDS faces.

\subsection{Computational Efficiency}
While LOBSTER demonstrates competitive results in all different scenarios,
its primary contribution lies in its computational efficiency. Table \ref{tab:150_setup_avg_time} compares the average runtime (in seconds) for a single active learning selection iteration across different strategies, datasets and models.

\begin{table}[h]
\centering
\caption{Average runtime (in seconds) for instance selection, for one active learning iteration, across different  strategies. Comparison across backbones and datasets.}
\label{tab:150_setup_avg_time}
\resizebox{\columnwidth}{!}{%
\begin{tabular}{llrrrr}
\toprule
 &  & \multicolumn{4}{c}{\textbf{Avg. Selection Time (s)}} \\ 
\cmidrule{3-6}
\textbf{Model} & \textbf{Dataset} & BAS & DUAL & IDDS & LOBSTER \\ 
\midrule
\multirow{3}{*}{\begin{tabular}{c} BART-base\end{tabular}}
 & AESLC & 46.1 & 38.7 & 0.02 & 0.6 \\
 & XSum & 66.6 & 63.7 & 0.4 & 0.8 \\
 & CNN/DM & 70.2 & 69.2 & 0.3 & 0.8 \\ 
\midrule
\multirow{3}{*}{\begin{tabular}{c}PEGASUS-large\end{tabular}}
 & AESLC & 221.1 & 113.6 & 0.01 & 1.0 \\
 & XSum & 390.0 & 267.6 & 0.4 & 1.5 \\
 & CNN/DM & 1064.2 & 699.4 & 0.4 & 1.6 \\ 
\bottomrule
\end{tabular}%
}
\end{table}

\paragraph{Scalability and Runtime Analysis.}
Strategies relying on autoregressive decoding (e.g., BAS, DUAL) become prohibitively expensive as model and dataset sizes increase. These approaches estimate uncertainty by generating multiple distinct summaries for each candidate document via multiple stochastic forward passes (Monte Carlo dropout). Because these passes must be repeated at each active learning iteration to reassess the unlabeled pool, the computational overhead grows considerably with larger datasets and model scales. These strategies exhibit poor scalability, becoming increasingly impractical as model architectures scale up or as generation complexity increases.

In contrast, LOBSTER adopts a deterministic approach that shifts the computational burden from the expansive unlabeled pool to the much smaller labeled set. After the training phase, we compute the per-example cross-entropy loss using teacher forcing, which allows parallelized forward passes that are significantly faster than autoregressive token-by-token generation. Furthermore, LOBSTER replaces repeated model inference with efficient vector similarity computations in the embedding space, avoiding the expensive autoregressive bottleneck entirely and ensuring scalability for large-scale document pools and high-parameter models.

As demonstrated in Table \ref{tab:150_setup_avg_time}, LOBSTER consistently and significantly outperforms both BAS and DUAL in terms of computational efficiency (we omit random sampling as its selection overhead is negligible). For example, with BART-Base on AESLC, LOBSTER reduces selection time to just 0.6 seconds, compared to 46.1 seconds for BAS and 38.7 seconds for DUAL. This efficiency gap widens as the model and document length increase; on CNN/DM with PEGASUS-Large, BAS and DUAL require 1064.2 and 699.4 seconds respectively, whereas LOBSTER completes the selection in only 1.6 seconds. This shows that LOBSTER achieves a speedup of 665$\times$ over BAS and 437$\times$ over DUAL in this configuration, effectively reducing the selection latency from nearly 18 minutes to near real-time (approaching the near-zero overhead of the model-agnostic IDDS) and enabling LOBSTER's viability for large-scale summarization applications. Crucially, LOBSTER achieves this near-zero latency without sacrificing the high summarization quality of uncertainty-based methods or suffering the cold-start instability of density-based baselines like IDDS.


\subsection{Ablation Studies}
\subsubsection{Impact of Loss-Based Hard Example Selection}
To investigate the contribution of the loss-based hard example selection mechanism (Stage 1), we replaced the high-loss example selection with a random selection strategy. Specifically, instead of selecting the top-$k$ examples with the highest generation loss from the labeled set $\mathcal{L}$, we randomly sampled $k$ examples as semantic anchors. The subsequent stages of LOBSTER, including density-based pre-filtering and semantic hardness projection, remained unchanged. This ablation evaluates whether selecting high-loss examples provides an advantage over random anchor selection. As illustrated in Figure~\ref{fig:ablation_results_grid_bertscore}, on CNN/DM, LOBSTER outperforms the random-anchor variant across both BART and PEGASUS, suggesting that high-loss examples provide more effective semantic anchors for this dataset. In contrast, the performance gap on AESLC and XSum is relatively small, although standard LOBSTER maintains a slight advantage. This can be attributed to the fact that CNN/DM contains longer documents and summaries, making loss-guided example selection more beneficial. Since similar trends were observed across both ROUGE and BERTScore metrics, we report the BERTScore curves, which provide a complementary semantic evaluation of the generated summaries.

\subsubsection{Ablation Study: Impact of IDDS}
To isolate the contribution of the density-based pre-filtering (Stage 2), we removed the representative candidate pool ($\mathcal{U}_{rep}$) and applied the semantic hardness projection (Stage 3) directly to the entire unlabeled pool ($\mathcal{U}$). We evaluated this ablation on AESLC and XSum to assess robustness across different summarization styles.

\paragraph{Visualization of Semantic Collapse}

To visualize the behavior of the selection mechanism (Stage 3) without filtering, we applied PCA to Sentence-BERT embeddings. Figure~\ref{fig:ablation_strip} plots selected samples (red) against the unlabeled pool (blue), revealing a critical failure mode: without IDDS, LOBSTER becomes concentrated in specific embedding regions. This reduces exploration of the data distribution and leads to redundant selections, particularly in datasets such as AESLC and XSum that contain many semantically similar examples. In contrast, pre-filtering with IDDS mitigates this issue by constructing a representative candidate pool ($U_{\text{rep}}$) that preserves broader coverage of the unlabeled distribution before semantic hardness projection. This diversity constraint prevents the selection process from repeatedly focusing on local clusters, enabling more informative use of the annotation budget.

\subsection{Data Efficiency Setup}
\label{subsec:data_efficiency_setup}

In preliminary experiments with a 150-sample budget, several configurations failed to reach 90\% of full-dataset performance. Since such a small annotation budget is often unrealistic in practical settings, we expanded the budget for these underperforming setups (\textit{BART/AESLC}, \textit{PEGASUS/AESLC}, and \textit{BART/XSum}) to 2,000 samples ($B=100$ over 20 iterations). Configurations already achieving the 90\% target (e.g., \textit{CNN/DM}) required no expansion.        Figure~\ref{fig:main_results_grid_data_efficiency_setup} and Appendix~\ref{sec:appendix_data_efficiency} detail the convergence rates and sample requirements needed to reach the 90\% performance threshold, alongside a comprehensive runtime analysis of the selection process.

\paragraph{The Competitiveness of Random Sampling}
Table~\ref{tab:data_efficiency_analysis} in Appendix~\ref{sec:appendix_data_efficiency} reveals that the performance gap between random sampling and active learning methods narrows as the annotation budget increases. In several configurations, particularly on AESLC and XSum, random sampling matches or surpasses sophisticated active learning strategies. For example, in the \textit{BART-base/XSum} setup, random sampling reaches the 90\% threshold using only 1,600 samples, outperforming all other approaches. IDDS fails to reach this threshold in half of the evaluated scenarios, while BAS succeeds on the \textit{PEGASUS-large/CNN/DM} setup at the cost of prohibitive computational latency. LOBSTER reaches the target in 5 out of 6 configurations, providing a balance between reliability and efficiency. This competitive behavior of random sampling at larger annotation budgets can be attributed to the relatively consistent structure of standard summarization benchmarks. As the annotation budget increases, larger uniformly sampled subsets are increasingly likely to capture the underlying document distribution, naturally reducing the advantage of more complex acquisition strategies. Consequently, active learning frameworks provide the greatest benefit under strict low-budget constraints, such as our primary 150-instance setting, where careful instance selection is critical for maximizing annotation efficiency.

\subsection{Comparison with Modern LLMs}
Given the increasing adoption of instruction-tuned large language models for summarization, we compare models fine-tuned using examples selected by LOBSTER (at a 150-instance annotation budget) against a zero-shot Llama-3-8B-Instruct baseline. As shown in Table~\ref{tab:llm_comparison} in Appendix~\ref{sec:appendix_data_efficiency}, LOBSTER remains highly competitive and, in several cases, outperforms the zero-shot LLM across both lexical (ROUGE) and semantic (BERTScore) metrics. These findings demonstrate that carefully selecting only 150 informative training examples through active learning is sufficient for a compact encoder-decoder model to achieve competitive summarization performance. Beyond accuracy, this approach offers important practical advantages, as such models typically require substantially less memory and lower inference latency than multi-billion-parameter autoregressive LLMs and, when deployed locally, avoid the recurring token costs associated with API-based LLM services.

\section{Conclusions and Future Work}
This work demonstrated that the cross-entropy loss of labeled examples can serve as an effective acquisition signal for active learning in abstractive summarization. By identifying the model's high-loss examples from the labeled set, our approach targets the specific areas where the model struggles. Building on this, we select the unlabeled ``semantic twins'' of those high-loss cases so the model is explicitly guided to correct its specific weaknesses. Furthermore, we showed that using a density-based pre-filter (IDDS) to find a representative candidate pool before final selection prevents semantic collapse. Extensive evaluation across three benchmark datasets and two backbone architectures confirms that LOBSTER matches or outperforms state-of-the-art baselines while remaining significantly faster. Finally, our data efficiency analysis indicates that random sampling remains surprisingly competitive under large annotation budgets. As the budget increases, random selection captures enough of the underlying data distribution to yield strong summarization performance. As future work, we plan to explore the applicability of LOBSTER to other sequence-to-sequence generation tasks beyond abstractive summarization, as well as its adaptation to modern LLMs. 

\section*{Limitations}
While our study demonstrates promising results, a few limitations remain. Our experiments are limited to English-language datasets (AESLC, XSum, and CNN/DailyMail), and therefore the effectiveness of LOBSTER for other languages remains unexplored. Languages with different morphology, syntax, or summarization styles may influence the behavior of embedding similarity and active learning selection strategies. Second, although we evaluate using both lexical (ROUGE) and semantic (BERTScore) metrics, automatic evaluation metrics may not fully capture aspects such as factual consistency, coherence, and overall summary quality. In addition, LOBSTER assumes that semantic similarity in the Sentence-BERT embedding space provides a useful proxy for transferring hardness from labeled to unlabeled examples. Although our empirical results support this design choice, investigating alternative retrieval spaces or jointly learned task-specific embeddings remains an important direction for future work. Finally, our experiments simulate human annotation by retrieving ground-truth summaries from existing datasets rather than collecting new human annotations, which may not fully reflect real-world annotation scenarios.

\section*{Ethical Considerations}
This work focuses on improving the efficiency of active learning for abstractive text summarization. The proposed method does not introduce new datasets, but it relies on pre-trained language models and existing benchmark corpora, which may contain biases present in their original training data. As a result, biases in the underlying models or datasets could influence the selection of instances during the active learning process and may propagate to downstream summarization systems. Future work could explore bias-aware acquisition strategies or incorporate fairness constraints when selecting samples for annotation.

\bibliography{custom}
\clearpage
\appendix
\FloatBarrier
\section*{Appendix}

\section{Dataset Statistics and Model Hyperparameters}
\label{sec:appendix_dataset_statistics_and_hyperparameters}
In this section, we provide additional details regarding the datasets and models used in our experiments. Table~\ref{tab:dataset_stats} reports dataset statistics, including the number of instances and the average lengths of source documents and summaries for both training and test splits. Table~\ref{tab:hyperparameters} provides the hyperparameters used for fine-tuning the backbone summarization models. 

\begin{table}[H]
    \centering
       \caption{Statistics for the datasets used in this work. We report the number of instances (\# Ins.) and the average length of source documents (Doc. len.) and summaries (Sum. len.) for both training and test splits.}
    \label{tab:dataset_stats}
    \resizebox{\columnwidth}{!}{%
        \begin{tabular}{llrrr}
        \toprule
        \textbf{Dataset} & \textbf{Subset} & \textbf{\# Ins.} & \textbf{Doc. len.} & \textbf{Sum. len.} \\
        \midrule
        \multirow{2}{*}{\textit{Gigaword}} 
            & Train & 200 & 39.6 & 11.2 \\
            & Test  & 1K  & 38.3 & 10.4 \\
        \midrule
        \multirow{2}{*}{\textit{AESLC}} 
            & Train & 14,436 & 115.3 & 4 \\
            & Test  & 1,906  & 114.6 & 4.1 \\
        \midrule
        \multirow{2}{*}{\textit{XSum}} 
            & Train & 204,045 & 388.6 & 23.1 \\
            & Test  & 1K  & 384.3 & 23 \\
        \midrule
        \multirow{2}{*}{\textit{CNN/DM}} 
            & Train & 287,113 & 692.4 & 51.6 \\
            & Test  & 1K  &676.1  & 55.2 \\
        \bottomrule
        \end{tabular}%
        }
    
\end{table}

\begin{table}[H]
\centering
\setlength{\tabcolsep}{5pt}
\caption{Hyperparameters used for fine-tuning the backbone summarization models.}
\label{tab:hyperparameters}
\resizebox{\columnwidth}{!}{%
\begin{tabular}{lcc}
\toprule
\textbf{Hyperparameter} & \textbf{BART-base} & \textbf{PEGASUS-large} \\
\midrule
Learning Rate & $2e^{-5}$ & $5e^{-5}$ \\
Batch Size (Train) & 16 & 8 \\
Batch Size (Eval) & 32 & 16 \\
Num. Epochs & 6 & 4 \\
Optimizer & AdamW & AdamW \\
Weight Decay & 0.028 & 0.03 \\
Warmup Ratio & 0.1 & 0.1 \\
Beam Size & 4 & 4 \\
Min. Train Steps & 350 & 200 \\
Grad. Accumulation & 1 & 1 \\
\bottomrule
\end{tabular}
}
\end{table}

In this work, we utilize standard public datasets (AESLC, XSum, CNN/DailyMail) and pre-trained models (e.g., BART, PEGASUS). These artifacts are established benchmarks within the NLP community and are used for academic benchmarking and summarization evaluation. All datasets and models are used in accordance with their respective open-source licenses and research usage guidelines. Because our core contribution focuses on an algorithmic active learning strategy rather than novel dataset curation, we relied on the standard versions and original preprocessing of these datasets. Consequently, we did not perform additional data collection. We also did not apply further filtering for personally identifiable information or offensive content beyond the preprocessing steps already applied by the original dataset creators. 

All experiments were conducted using publicly available implementations of the selected backbone models through the HuggingFace Transformers library. Training and evaluation procedures follow standard fine-tuning practices for abstractive summarization models. Hyperparameters were selected based on preliminary tuning experiments and remain consistent across all active learning strategies to ensure a fair comparison.

To promote reproducibility, all code created for this study, including the full experimental pipeline and the implementation of the proposed active learning method, is provided anonymously solely for the intended purpose of peer-review.

\section{Summarization Performance}
\label{sec:appendix_summarization_performance}
Table \ref{tab:main_results_combined} reports summarization performance of all methods under a 150-instance annotation budget across AESLC, XSum, and CNN/DailyMail using ROUGE and BERTScore. Statistical significance tests between LOBSTER and baselines are reported in Table \ref{tab:statistical_tests}.

\begin{table*}[hbt!]
\centering
\caption{Summarization performance (ROUGE and rescaled BERTScore) with a 150-instance annotation budget. Results are reported as mean $\pm$ standard deviation. The best performance for each metric and dataset is highlighted in bold.}
\label{tab:main_results_combined}
\begin{tabular}{ll cccc}
\toprule
\textbf{Model} & \textbf{Method} & \textbf{ROUGE-1} & \textbf{ROUGE-2} & \textbf{ROUGE-L} & \textbf{BERTScore} \\
\midrule
\multicolumn{6}{c}{\textbf{Dataset: AESLC}} \\
\midrule
\multirow{5}{*}{BART-base} 
 & Random  & 26.6$\pm$0.65 & 13.2$\pm$0.65 & 26.0$\pm$0.63 & 13.51$\pm$0.57 \\
 & BAS     & 26.5$\pm$0.17 & 13.1$\pm$0.31 & 25.8$\pm$0.24 & \textbf{13.63}$\pm$0.99 \\
 & IDDS    & 27.0$\pm$0.32 & 12.5$\pm$0.32 & 25.4$\pm$0.31 & 13.03$\pm$0.19 \\
 & DUAL    & 26.9$\pm$0.86 & 13.2$\pm$0.70 & 26.0$\pm$0.85 & 12.95$\pm$0.14 \\
 & LOBSTER & \textbf{27.1}$\pm$0.88 & \textbf{13.4}$\pm$0.45 & \textbf{26.3}$\pm$0.80 & 12.96$\pm$0.65 \\
\midrule
\multirow{5}{*}{PEGASUS-large} 
 & Random  & 21.1$\pm$0.54 & 9.9$\pm$0.48 & 20.0$\pm$0.55 & 6.14$\pm$0.68 \\
 & BAS     & 19.6$\pm$1.34 & 9.0$\pm$0.86 & 18.9$\pm$1.36 & 4.82$\pm$1.36 \\
 & IDDS    & \textbf{22.6}$\pm$0.42 & \textbf{11.0}$\pm$0.34 & \textbf{21.5}$\pm$0.47 & \textbf{7.70}$\pm$0.59 \\
 & DUAL    & 21.2$\pm$1.15 & 10.1$\pm$0.47 & 20.5$\pm$1.08 & 6.29$\pm$0.90 \\
 & LOBSTER & 22.1$\pm$1.01 & 10.9$\pm$0.61 & \textbf{21.5}$\pm$0.94 & 7.66$\pm$0.55 \\

\midrule
\multicolumn{6}{c}{\textbf{Dataset: XSum}} \\
\midrule
\multirow{5}{*}{BART-base} 
 & Random  & 32.6$\pm$0.36 & 10.7$\pm$0.38 & 25.3$\pm$0.38 & 37.27$\pm$0.35 \\
 & BAS     & 32.3$\pm$0.15 & 10.6$\pm$0.14 & 25.1$\pm$0.23 & 36.93$\pm$0.30 \\
 & IDDS    & 32.6$\pm$0.14 & 10.6$\pm$0.10 & 25.0$\pm$0.17 & 36.71$\pm$0.15 \\
 & DUAL    & \textbf{32.9}$\pm$0.17 & \textbf{10.8}$\pm$0.15 & \textbf{25.6}$\pm$0.14 & \textbf{37.42}$\pm$0.28 \\
 & LOBSTER & 32.5$\pm$0.21 & 10.6$\pm$0.21 & 25.3$\pm$0.28 & 37.00$\pm$0.22 \\
\midrule
\multirow{5}{*}{PEGASUS-large} 
 & Random  & 43.4$\pm$0.37 & 20.1$\pm$0.31 & 34.9$\pm$0.50 & \textbf{48.00}$\pm$0.37 \\
 & BAS     & 43.3$\pm$0.30 & 20.2$\pm$0.33 & 34.9$\pm$0.36 & 47.71$\pm$0.40 \\
 & IDDS    & 42.8$\pm$0.13 & 19.6$\pm$0.12 & 34.2$\pm$0.11 & 47.28$\pm$0.24 \\
 & DUAL    & 43.3$\pm$0.24 & \textbf{20.3}$\pm$0.30 & \textbf{35.0}$\pm$0.42 & 47.74$\pm$0.38 \\
 & LOBSTER & \textbf{43.5}$\pm$0.49 & \textbf{20.3}$\pm$0.49 & \textbf{35.0}$\pm$0.47 & 47.72$\pm$0.35 \\

\midrule
\multicolumn{6}{c}{\textbf{Dataset: CNN/DM}} \\
\midrule
\multirow{5}{*}{BART-base} 
 & Random  & \textbf{39.5}$\pm$0.24 & 17.0$\pm$0.18 & 26.3$\pm$0.18 & 29.35$\pm$0.24 \\
 & BAS     & \textbf{39.5}$\pm$0.22 & 17.0$\pm$0.19 & \textbf{26.6}$\pm$0.16 & 29.42$\pm$0.18 \\
 & IDDS    & 39.2$\pm$0.29 & 16.5$\pm$0.18 & 26.2$\pm$0.14 & 29.40$\pm$0.18 \\
 & DUAL    & 39.4$\pm$0.66 & \textbf{17.1}$\pm$0.37 & \textbf{26.6}$\pm$0.32 & \textbf{29.44}$\pm$0.85 \\
 & LOBSTER & 39.4$\pm$0.58 & 17.0$\pm$0.48 & 26.4$\pm$0.39 & 29.37$\pm$0.20 \\
\midrule
\multirow{5}{*}{PEGASUS-large} 
 & Random  & 39.2$\pm$0.24 & 17.0$\pm$0.15 & 26.3$\pm$0.20 & 26.40$\pm$0.15 \\
 & BAS     & \textbf{39.4}$\pm$0.32 & \textbf{17.1}$\pm$0.20 & 26.2$\pm$0.17 & 26.17$\pm$0.32 \\
 & IDDS    & 38.8$\pm$0.33 & 16.8$\pm$0.20 & 26.2$\pm$0.20 & 25.37$\pm$0.30 \\
 & DUAL    & 39.2$\pm$0.37 & \textbf{17.1}$\pm$0.18 & \textbf{26.4}$\pm$0.21 & \textbf{26.49}$\pm$0.33 \\
 & LOBSTER & 39.2$\pm$0.43 & 16.9$\pm$0.22 & 26.2$\pm$0.20 & 25.96$\pm$0.48 \\
\bottomrule
\end{tabular}
\end{table*}


\begin{table*}[ht]
\centering
\small
\caption{Paired permutation test p-values comparing LOBSTER against baseline methods at the 150-instance annotation budget for ROUGE-1 and BERTScore. An asterisk (*) indicates a statistically significant difference ($p < 0.05$). For significant results, arrows indicate whether LOBSTER scored significantly higher ($\uparrow$) or lower ($\downarrow$) than the baseline.}
\setlength{\tabcolsep}{6pt}
\begin{tabular}{ll|cccc}
\toprule
 & & \multicolumn{4}{c}{\textbf{LOBSTER vs. Baseline ($p$-value)}} \\
\textbf{Dataset} & \textbf{Model} & \textbf{DUAL} & \textbf{Random} & \textbf{IDDS} & \textbf{BAS} \\
\midrule
\multicolumn{6}{l}{\textit{\textbf{ROUGE-1}}} \\
\multirow{2}{*}{AESLC} 
 & BART-base     & 0.1681 & 0.4250 & 0.0083$^*\uparrow$ & 0.0203$^*\uparrow$ \\
 & PEGASUS-large & $<$0.001$^*\uparrow$ & $<$0.001$^*\uparrow$ & 0.5113 & $<$0.001$^*\uparrow$ \\
\midrule
\multirow{2}{*}{XSum} 
 & BART-base     & 0.0224$^*\downarrow$ & 0.1649 & 0.6326 & 0.2942 \\
 & PEGASUS-large & 0.2192 & 0.8924 & $<$0.001$^*\uparrow$ & 0.2532 \\
\midrule
\multirow{2}{*}{CNN/DM} 
 & BART-base     & 0.1807 & 0.0722 & 0.8836 & 0.0106$^*\downarrow$ \\
 & PEGASUS-large & $<$0.001$^*\downarrow$ & $<$0.001$^*\downarrow$ & 0.2663 & $<$0.001$^*\downarrow$ \\
\toprule
\multicolumn{6}{l}{\textit{\textbf{BERTScore}}} \\
\multirow{2}{*}{AESLC} 
 & BART-base     & 0.3666 & 0.0074$^*\downarrow$ & 0.3685 & 0.0013$^*\downarrow$ \\
 & PEGASUS-large & $<$0.001$^*\uparrow$ & $<$0.001$^*\uparrow$ & 0.9037 & $<$0.001$^*\uparrow$ \\
\midrule
\multirow{2}{*}{XSum} 
 & BART-base     & 0.0150$^*\downarrow$ & 0.1250 & 0.2651 & 0.7348 \\
 & PEGASUS-large & 0.9361 & 0.0946 & 0.0338$^*\uparrow$ & 0.9475 \\
\midrule
\multirow{2}{*}{CNN/DM} 
 & BART-base     & $<$0.001$^*\downarrow$ & $<$0.001$^*\uparrow$ & $<$0.001$^*\downarrow$ & $<$0.001$^*\downarrow$ \\
 & PEGASUS-large & $<$0.001$^*\downarrow$ & $<$0.001$^*\downarrow$ & 0.6113 & 0.0100$^*\downarrow$ \\
\bottomrule
\end{tabular}
\label{tab:statistical_tests}
\end{table*}

\begin{table*}[ht]
\centering
\small
\caption{Performance comparison of models fine-tuned using examples selected by LOBSTER (at a 150-instance annotation budget) against a zero-shot Llama-3-8B-Instruct baseline. Bold values indicate the best performance for each metric within each dataset.}
\setlength{\tabcolsep}{6pt}
\begin{tabular}{llcc}
\toprule
\textbf{Dataset} & \textbf{Setup} & \textbf{ROUGE (1 / 2 / L)} & \textbf{BERTScore} \\
\midrule
\multirow{3}{*}{AESLC} 
 & Llama-3-8B-Instruct (Zero-shot) & \textbf{27.16} / 12.06 / 25.22 & 11.61 \\
 & BART-base + LOBSTER             & 27.10 / \textbf{13.40} / \textbf{26.30} & \textbf{12.96} \\
 & PEGASUS-large + LOBSTER         & 22.10 / 10.90 / 21.50 & \phantom{0}7.66 \\
\midrule
\multirow{3}{*}{XSum} 
 & Llama-3-8B-Instruct (Zero-shot) & 29.41 / \phantom{0}8.40 / 21.20 & 29.13 \\
 & BART-base + LOBSTER             & 32.50 / 10.60 / 25.30 & 37.00 \\
 & PEGASUS-large + LOBSTER         & \textbf{43.50} / \textbf{20.30} / \textbf{35.00} & \textbf{47.72} \\
\midrule
\multirow{3}{*}{CNN/DM} 
 & Llama-3-8B-Instruct (Zero-shot) & 39.39 / 15.61 / 24.72 & 26.28 \\
 & BART-base + LOBSTER             & \textbf{39.40} / \textbf{17.00} / \textbf{26.40} & \textbf{29.37} \\
 & PEGASUS-large + LOBSTER         & 39.20 / 16.90 / 26.20 & 25.96 \\
\bottomrule
\end{tabular}
\label{tab:llm_comparison}
\end{table*}

\section{Detailed Data Efficiency Setup Results}
\label{sec:appendix_data_efficiency}
In this section, we provide additional results for the data efficiency analysis discussed in Section~\ref{subsec:data_efficiency_setup}. 
Table~\ref{tab:data_efficiency_analysis} reports the number of annotated samples required for each method to reach 90\% of the full dataset performance across different datasets and backbone models. Additionally, Table~\ref{tab:data_efficiency_setup_avg_time} presents the average runtime required for instance selection during a single active learning iteration, highlighting the computational efficiency of the compared strategies.

These results are consistent with the trends discussed in the main text. In particular, Table~\ref{tab:data_efficiency_analysis} highlights the strong competitiveness of random sampling under larger annotation budgets, while LOBSTER also achieves strong and consistent performance across most setups. At the same time, LOBSTER provides a balance between performance reliability and computational efficiency compared to uncertainty-based approaches, as shown in Table~\ref{tab:data_efficiency_setup_avg_time}.

\begin{table*}[t] 
\centering
\caption{Data Efficiency Analysis: Number of annotated samples required to reach 90\% of the full-dataset performance. Bold values indicate the winning strategy for each configuration, and a dash (---) indicates that the configuration failed to reach the 90\% target.}
\label{tab:data_efficiency_analysis}
\begin{tabular}{llccccc}
\toprule
\textbf{Model} & \textbf{Dataset} & \multicolumn{5}{c}{\textbf{Active Learning}} \\ 
\cmidrule(lr){3-7}
& & \textbf{LOBSTER} & \textbf{DUAL} & \textbf{BAS} & \textbf{IDDS} & \textbf{Rand.} \\ 
\midrule
\multirow{3}{*}{BART-base} 
& AESLC & 1900 & 1900 & \textbf{1300} & --- & 2000 \\
& XSum & 2000 & 1700 & 1700 & --- & \textbf{1600} \\
& CNN/DM & \textbf{50} & 60 & 110 & 100 & 100 \\
\midrule
\multirow{3}{*}{\shortstack[l]{PEGASUS-large}} 
& AESLC & \textbf{1800} & \textbf{1800} & 2000 & 2000 & \textbf{1800} \\
& XSum & \textbf{80} & 90 & \textbf{80} & 100 & \textbf{80} \\
& CNN/DM & --- & --- & \textbf{150} & --- & --- \\
\bottomrule
\end{tabular}
\end{table*}

\begin{table*}[ht]
\centering
\caption{Data Efficiency Analysis: Average runtime (in seconds) for instance selection, for one active learning iteration, across different strategies. Comparison across backbones and datasets.}
\label{tab:data_efficiency_setup_avg_time}
\begin{tabular}{llrrrr}
\toprule
 &  & \multicolumn{4}{c}{\textbf{Avg Selection time (s)}} \\ 
\cmidrule{3-6}
\textbf{Model} & \textbf{Dataset} & BAS & DUAL & IDDS & LOBSTER \\ 
\midrule
\multirow{3}{*}{\begin{tabular}{c}BART-base\end{tabular}}
 & AESLC & 61.8 & 366.2 & 0.3 & 3.9 \\
 & XSum & 96.1 & 560.5 & 2.8 & 12.1 \\
 & CNN/DM & 70.2 & 69.2 & 0.3 & 0.8 \\ 
\midrule
\multirow{3}{*}{\begin{tabular}{c}PEGASUS-large\end{tabular}}
 & AESLC & 164.6 & 618.1 & 0.3 & 12.8 \\
 & XSum & 390.0 & 267.6 & 0.4 & 1.5 \\
 & CNN/DM & 1064.2 & 699.4 & 0.4 & 1.6 \\ 
\bottomrule
\end{tabular}%
\end{table*}

\begin{figure*}[t]
    \centering
    \renewcommand{\thesubfigure}{\Alph{subfigure}}
    
    \begin{subfigure}[t]{0.32\textwidth}
        \centering
        \includegraphics[width=\linewidth]{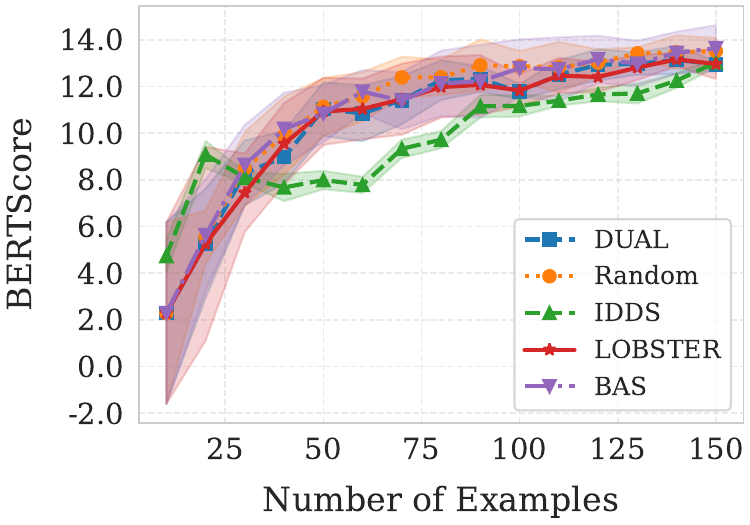}
        \caption{BART on AESLC}
        \label{fig:bart_aeslc}
    \end{subfigure}
    \hfill
    \begin{subfigure}[t]{0.32\textwidth}
        \centering
        \includegraphics[width=\linewidth]{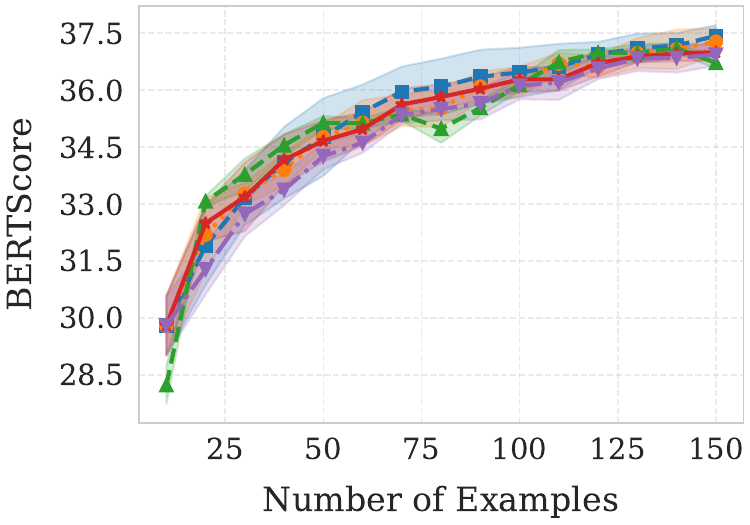}
        \caption{BART on XSum}
        \label{fig:bart_xsum}
    \end{subfigure}
    \hfill
    \begin{subfigure}[t]{0.32\textwidth}
        \centering
        \includegraphics[width=\linewidth]{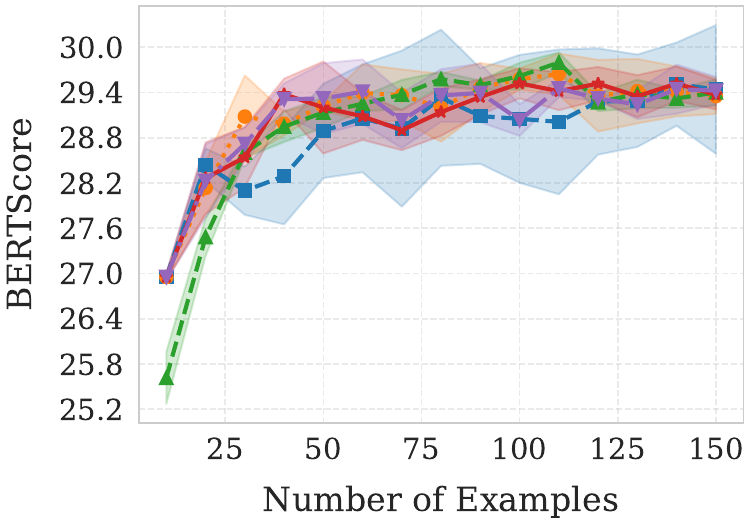}
        \caption{BART on CNN/DM}
        \label{fig:bart_cnn}
    \end{subfigure}

    \vspace{0.3cm} 

    \begin{subfigure}[t]{0.32\textwidth}
        \centering
        \includegraphics[width=\linewidth]{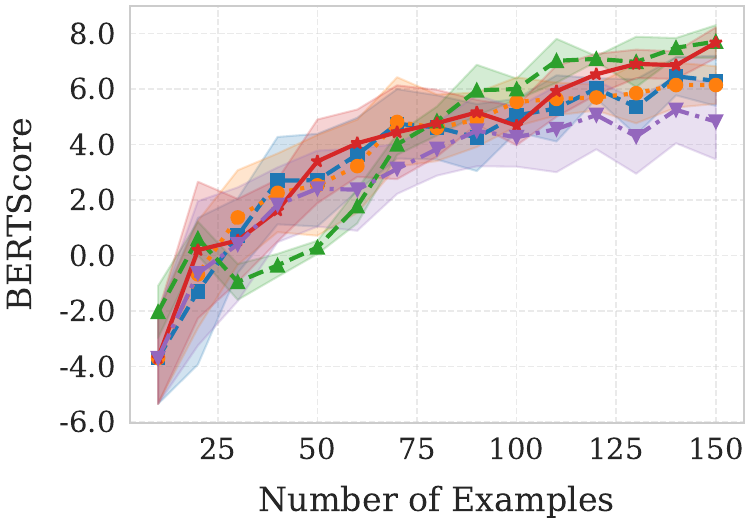}
        \caption{PEGASUS on AESLC}
        \label{fig:pegasus_aeslc}
    \end{subfigure}
    \hfill
    \begin{subfigure}[t]{0.32\textwidth}
        \centering
        \includegraphics[width=\linewidth]{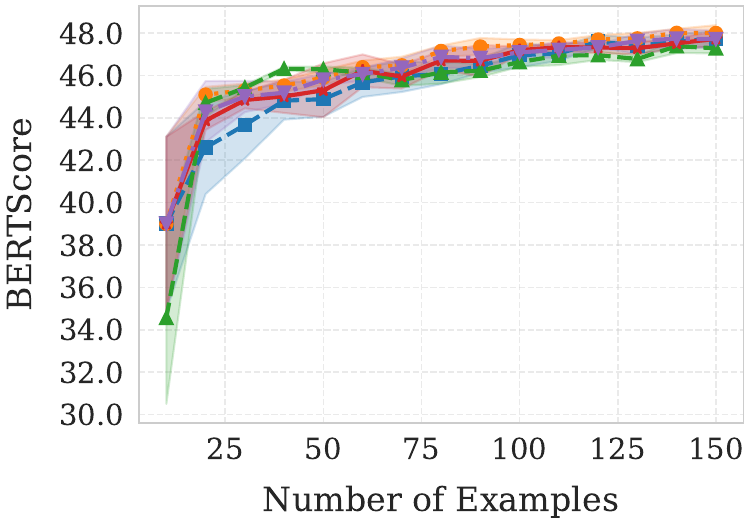}
        \caption{PEGASUS on XSum}
        \label{fig:pegasus_xsum}
    \end{subfigure}
    \hfill
    \begin{subfigure}[t]{0.32\textwidth}
        \centering
        \includegraphics[width=\linewidth]{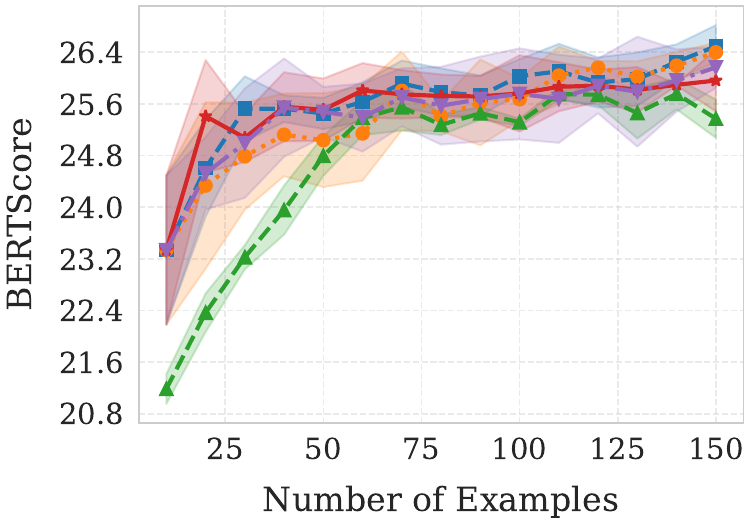}
        \caption{PEGASUS on CNN/DM}
        \label{fig:pegasus_cnn}
    \end{subfigure}

    \caption{Number of labeled instances vs. rescaled BERTScore. Comparison of active strategies using BART and PEGASUS backbones on AESLC, XSum, and CNN/DM datasets.}
    \label{fig:main_results_grid_bertscore}
\end{figure*}

\begin{figure*}[t]
    \centering
    \renewcommand{\thesubfigure}{\Alph{subfigure}}
    
    \begin{subfigure}[t]{0.32\textwidth}
        \centering
        \includegraphics[width=\linewidth]{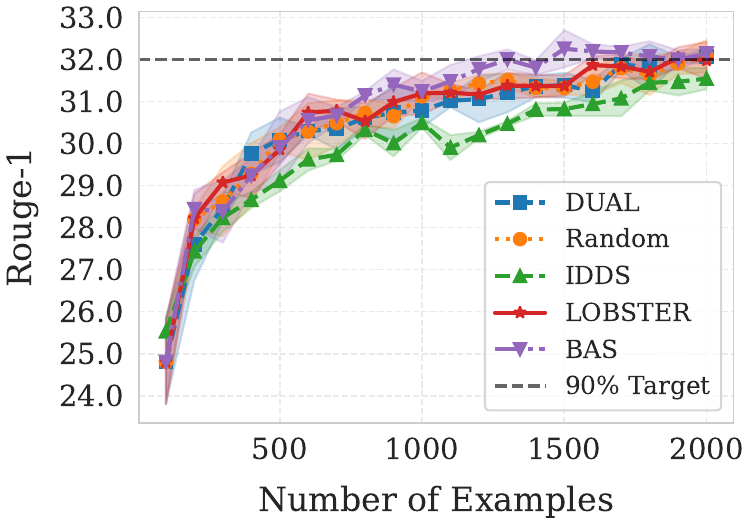}
        \caption{BART on AESLC}
        \label{fig:bart_aeslc}
    \end{subfigure}
    \hfill
    \begin{subfigure}[t]{0.32\textwidth}
        \centering
        \includegraphics[width=\linewidth]{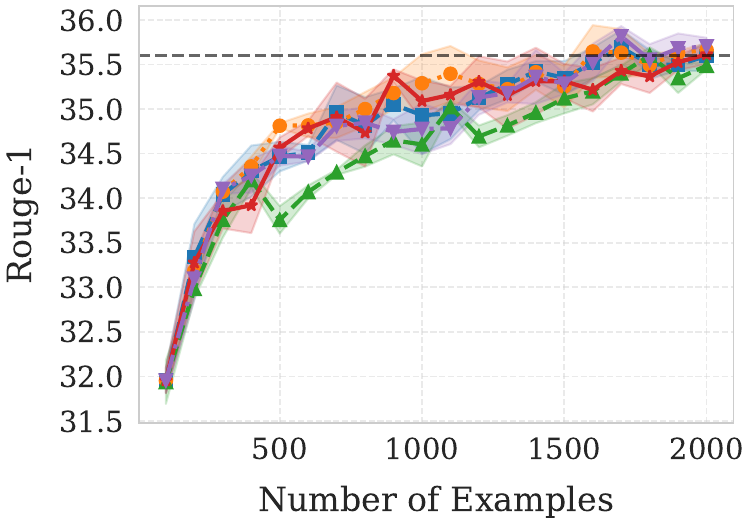}
        \caption{BART on XSum}
        \label{fig:bart_xsum}
    \end{subfigure}
    \hfill
    \begin{subfigure}[t]{0.32\textwidth}
        \centering
        \includegraphics[width=\linewidth]{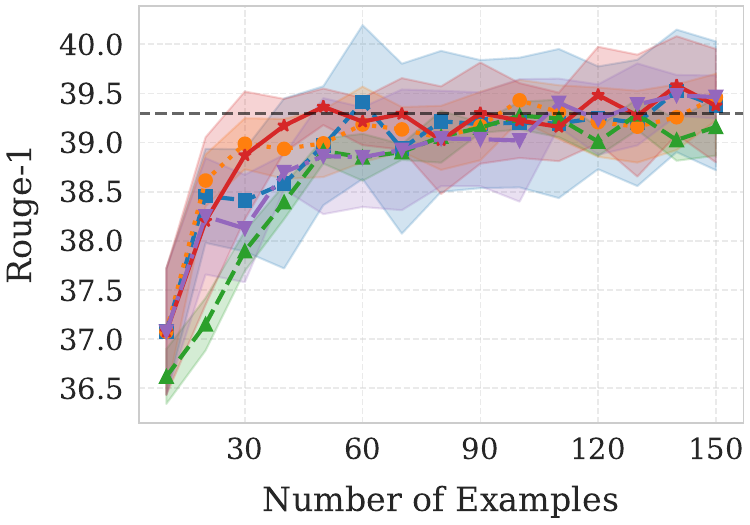}
        \caption{BART on CNN/DM}
        \label{fig:bart_cnn}
    \end{subfigure}

    \vspace{0.3cm} 

    \begin{subfigure}[t]{0.32\textwidth}
        \centering
        \includegraphics[width=\linewidth]{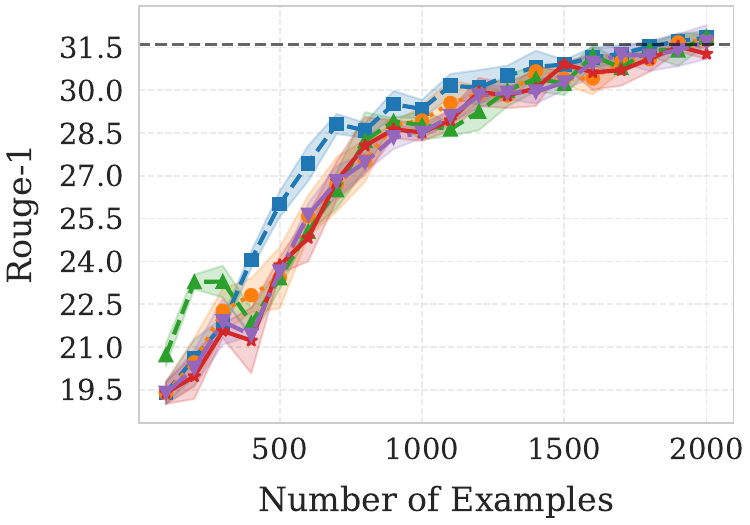}
        \caption{PEGASUS on AESLC}
        \label{fig:pegasus_aeslc}
    \end{subfigure}
    \hfill
    \begin{subfigure}[t]{0.32\textwidth}
        \centering
        \includegraphics[width=\linewidth]{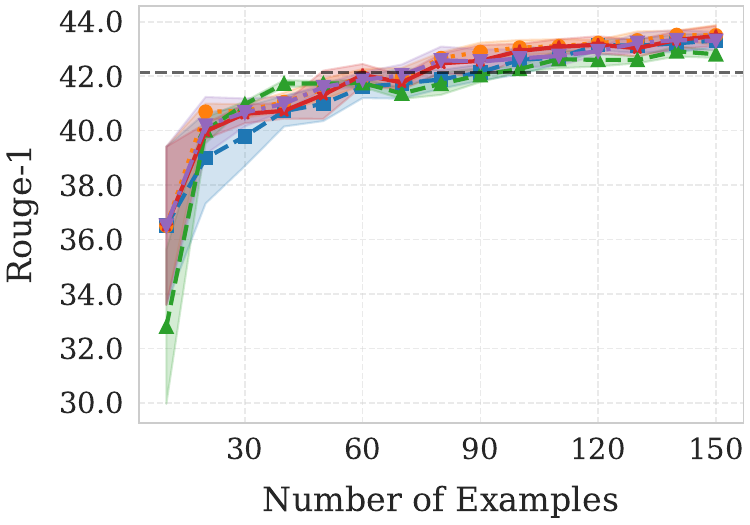}
        \caption{PEGASUS on XSum}
        \label{fig:pegasus_xsum}
    \end{subfigure}
    \hfill
    \begin{subfigure}[t]{0.32\textwidth}
        \centering
        \includegraphics[width=\linewidth]{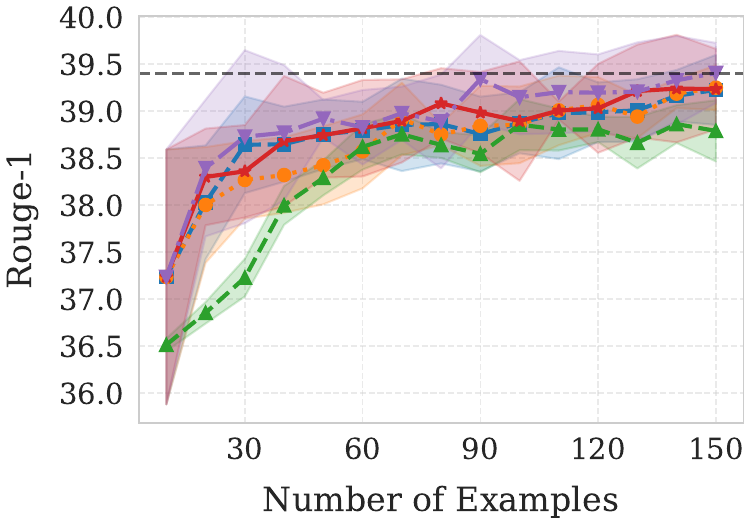}
        \caption{PEGASUS on CNN/DM}
        \label{fig:pegasus_cnn}
    \end{subfigure}

    \caption{Number of labeled instances vs. ROUGE-1. Comparison of active learning strategies using BART and PEGASUS backbones on AESLC, XSum, and CNN/DM datasets. The dashed horizontal line in each subplot denotes the 90\% performance threshold of the full-dataset baseline.}
    \label{fig:main_results_grid_data_efficiency_setup}
\end{figure*}

\begin{figure*}[t]
    \centering
    \renewcommand{\thesubfigure}{\Alph{subfigure}}
    
    \begin{subfigure}[t]{0.32\textwidth}
        \centering
        \includegraphics[width=\linewidth]{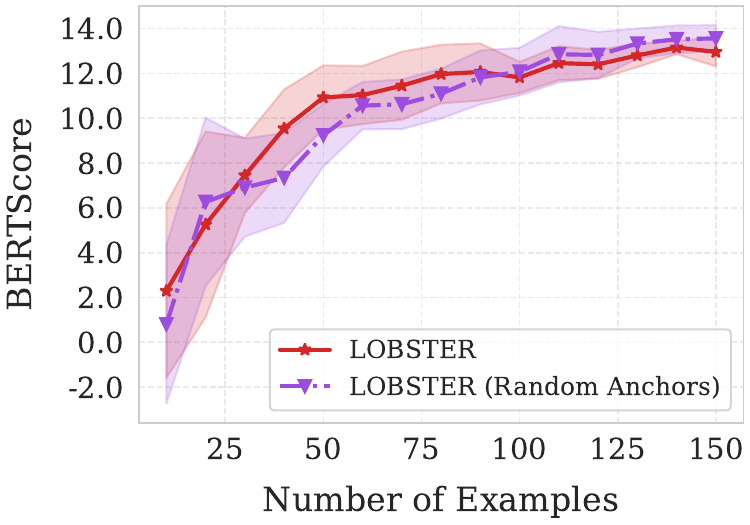}
        \caption{BART on AESLC}
        \label{fig:bart_aeslc}
    \end{subfigure}
    \hfill
    \begin{subfigure}[t]{0.32\textwidth}
        \centering
        \includegraphics[width=\linewidth]{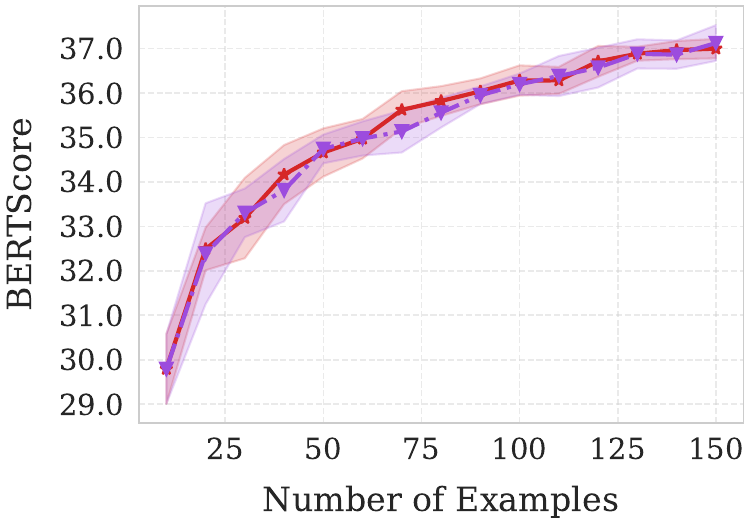}
        \caption{BART on XSum}
        \label{fig:bart_xsum}
    \end{subfigure}
    \hfill
    \begin{subfigure}[t]{0.32\textwidth}
        \centering
        \includegraphics[width=\linewidth]{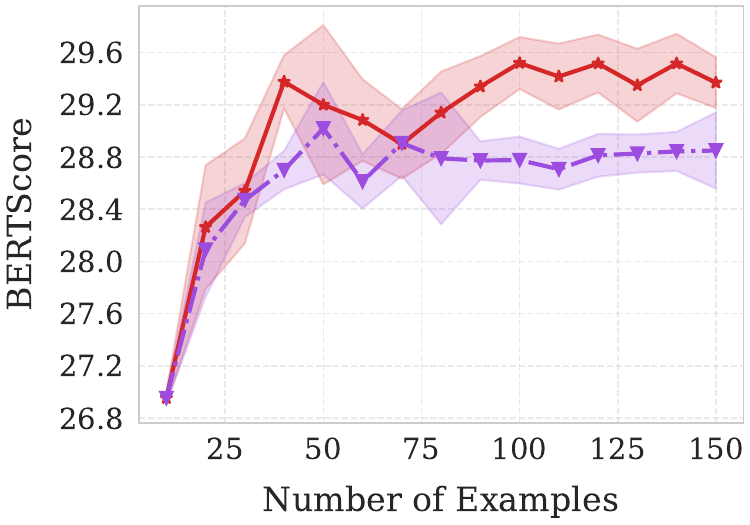}
        \caption{BART on CNN/DM}
        \label{fig:bart_cnn}
    \end{subfigure}

    \vspace{0.3cm} 

    \begin{subfigure}[t]{0.32\textwidth}
        \centering
        \includegraphics[width=\linewidth]{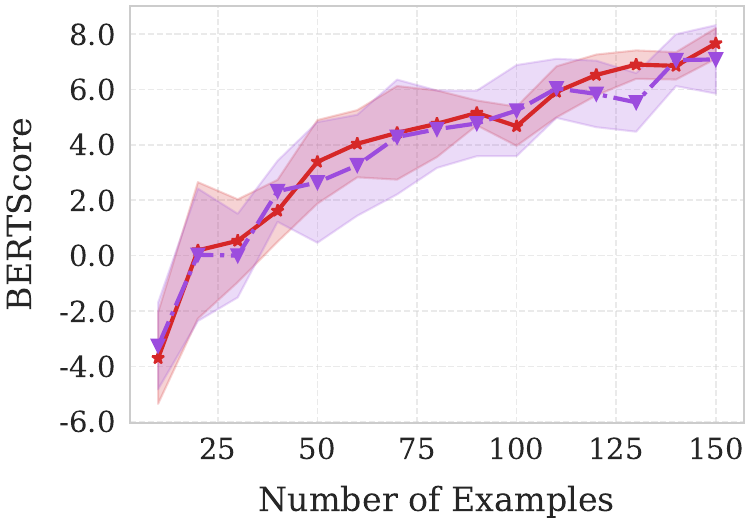}
        \caption{PEGASUS on AESLC}
        \label{fig:pegasus_aeslc}
    \end{subfigure}
    \hfill
    \begin{subfigure}[t]{0.32\textwidth}
        \centering
        \includegraphics[width=\linewidth]{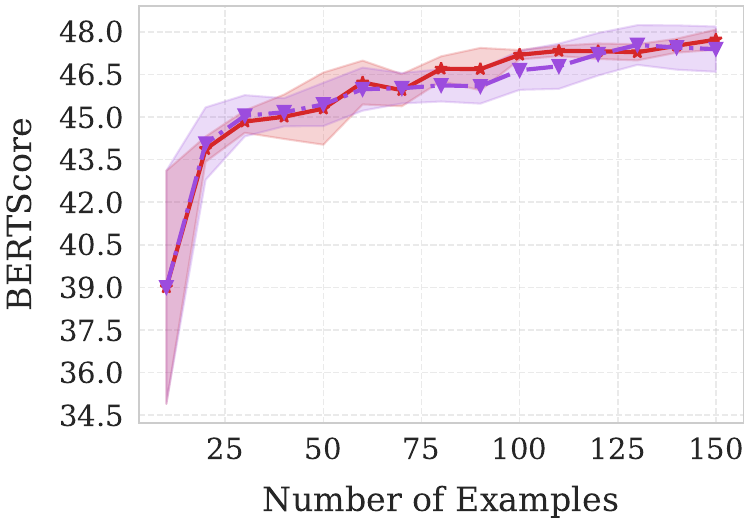}
        \caption{PEGASUS on XSum}
        \label{fig:pegasus_xsum}
    \end{subfigure}
    \hfill
    \begin{subfigure}[t]{0.32\textwidth}
        \centering
        \includegraphics[width=\linewidth]{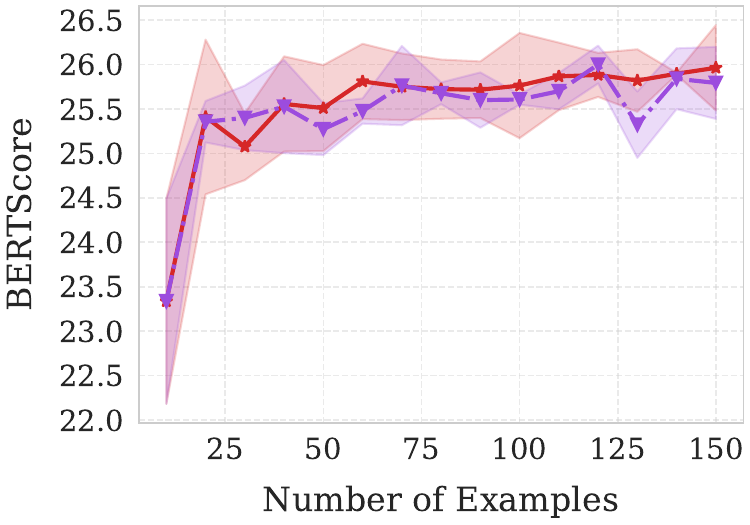}
        \caption{PEGASUS on CNN/DM}
        \label{fig:pegasus_cnn}
    \end{subfigure}

    \caption{Number of labeled instances vs. BERTScore. Comparison of LOBSTER and the LOBSTER (Random Anchors) ablation using BART and PEGASUS backbones on AESLC, XSum, and CNN/DM datasets.}
    \label{fig:ablation_results_grid_bertscore}
\end{figure*}

\begin{figure*}[t]
    \centering
    \begin{subfigure}[b]{0.24\textwidth}
        \centering
        {\footnotesize ROUGE-1: 25.3} \\[.1cm] 
        \includegraphics[width=\textwidth]{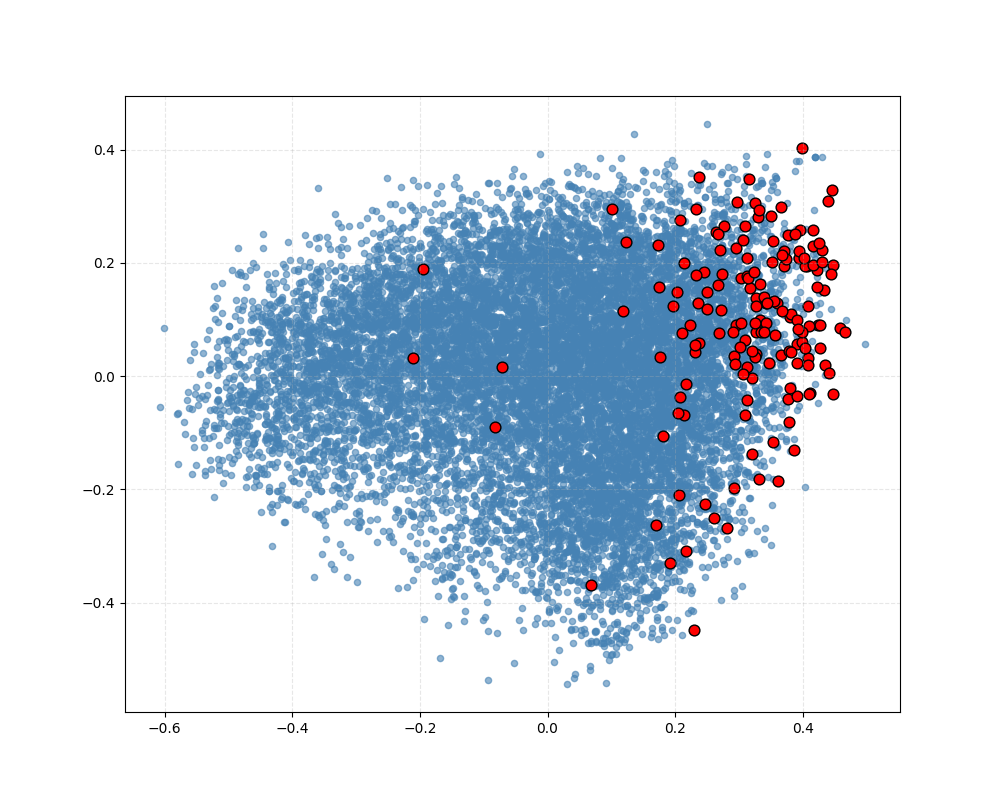}
        \caption{AESLC, without IDDS}
        \label{fig:aeslc_no_idds}
    \end{subfigure}
    \hfill
    \begin{subfigure}[b]{0.24\textwidth}
        \centering
        {\footnotesize ROUGE-1: 27.1 \\[.1cm] }
        \includegraphics[width=\textwidth]{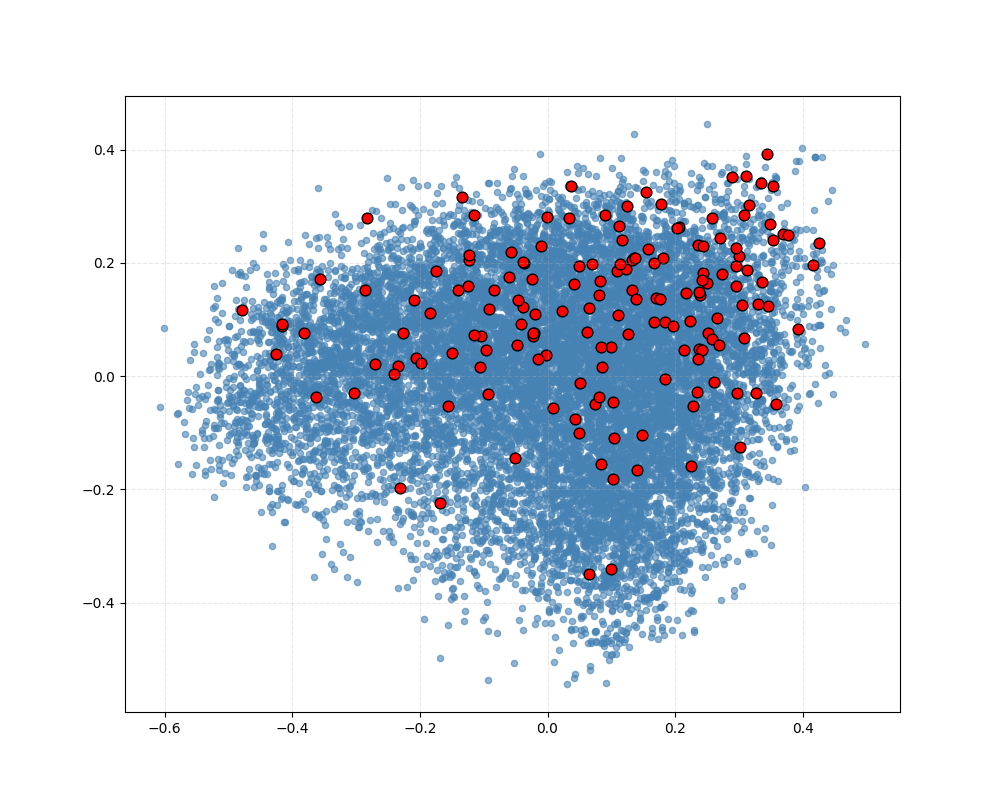}
        \caption{AESLC, with IDDS}
        \label{fig:aeslc_with_idds}
    \end{subfigure}
    \hfill
    \begin{subfigure}[b]{0.24\textwidth}
        \centering
        {\footnotesize ROUGE-1: 30.6 \\[.1cm]} 
        \includegraphics[width=\textwidth]{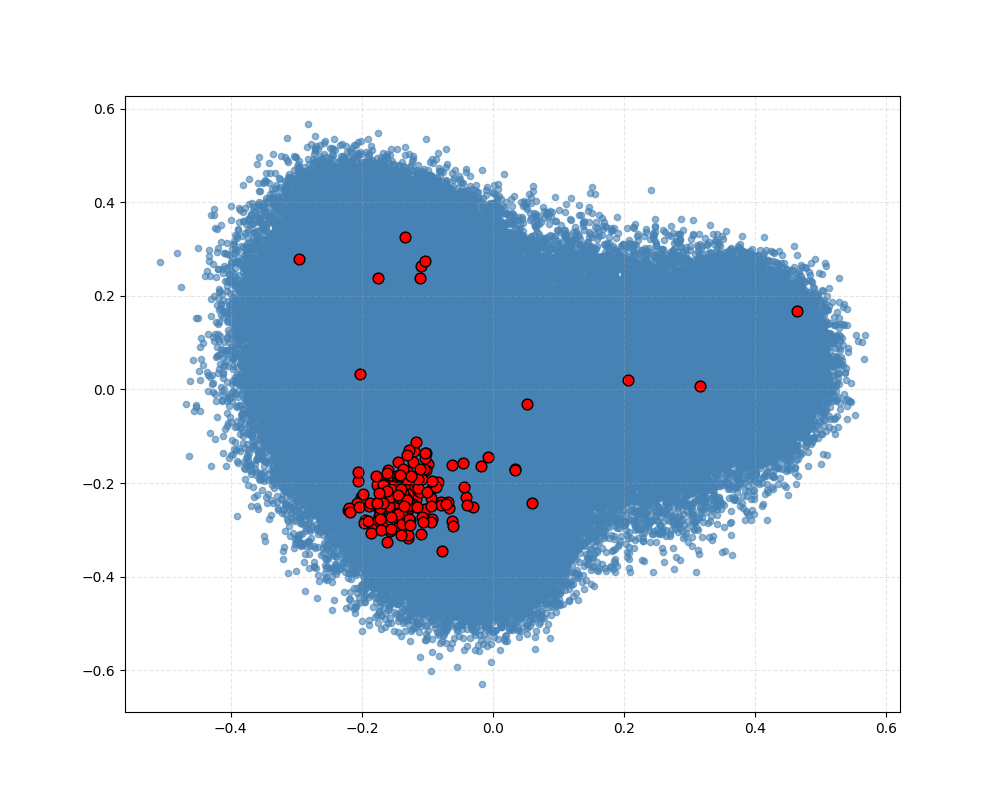}
        \caption{XSum, without IDDS}
        \label{fig:xsum_no_idds}
    \end{subfigure}
    \hfill
    \begin{subfigure}[b]{0.24\textwidth}
        \centering
        {\footnotesize ROUGE-1: 32.7 \\[.1cm]}
        \includegraphics[width=\textwidth]{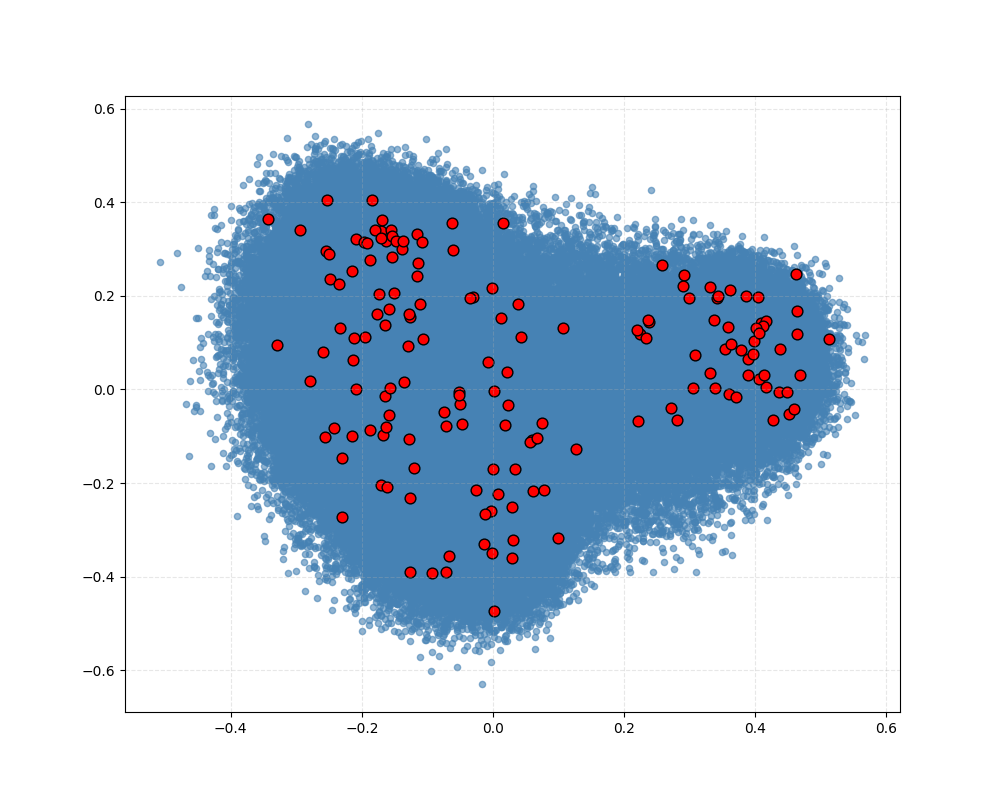}
        \caption{XSum, with IDDS}
        \label{fig:xsum_with_idds}
    \end{subfigure}

    \caption{Impact of IDDS on Selection Diversity (PCA visualization of selected instances). Comparison of the proposed LOBSTER method with IDDS versus the ablation without IDDS. The IDDS module prevents semantic collapse by filtering redundant candidates, leading to higher ROUGE-1 scores.}
    \label{fig:ablation_strip}
\end{figure*}
\end{document}